\documentclass[11pt]{article}

\usepackage[preprint]{acl}

\usepackage{times}
\usepackage{latexsym}
\usepackage[T1]{fontenc}
\usepackage[utf8]{inputenc}
\usepackage{microtype}
\usepackage{inconsolata}

\usepackage{amsmath}
\usepackage{graphicx}
\usepackage{url}
\usepackage{booktabs}
\usepackage{multirow}
\usepackage{algorithm}
\usepackage{algorithmic}
\usepackage{subcaption}
\usepackage{xcolor}

\title{Rethinking Literature Search Evaluation: \\
Deep Research Helps, and Human Citation Lists Are Not a Ground Truth}

\author{Gaurav Sahu$^{1,2}$ , Laurent Charlin$^{1,2,4}$, Christopher Pal$^{1,3,4,5,6}$  \\
  $^{1}$Mila -- Quebec AI Institute \quad $^{2}$HEC Montréal \\ $^{3}$ServiceNow Research
 \quad $^{4}$Canada CIFAR AI Chair \\ $^{5}$Université de Montréal \quad $^{6}$Polytechnique Montréal \\}

\begin{document}
\maketitle

\begin{abstract}
We study large-scale literature search from two complementary angles: improving the retrieval pipeline, and stress-testing the human reference list as an evaluation target. First, we implement a \emph{Deep Research} pipeline that processes the full query paper and expands the retrieved results breadth-first along their bibliographies, and show that it substantially outperforms vanilla API-only search, raising recall on \textsc{RollingEval-Jun25} (a 250-paper literature-search benchmark) from below 20\% to above 80\%.
Second, we use a neutral LLM-as-a-judge to determine if human references are sound ground truth for the task. We find significant limitations: only 51\% of human citations are judged moderately relevant or higher, against 86--88\% for the strongest AI-based re-rankers. We study this gap on the OpenAlex co-authorship graph, finding that humans are 2.5$\times$ more likely than the best AI re-rankers to cite a direct collaborator. Together, our results argue against single-axis literature-search evaluation: recall, topical-relevance scoring, ranked-list diversity, and a co-authorship-distance diagnostic each measure complementary properties of citation quality and should be reported jointly.
\end{abstract}

\section{Introduction}
Literature search remains a foundational step of scientific work, yet it is hard to evaluate well. The community typically reports precision and recall against the papers that the query paper itself cites \citep{ajith2024litsearch,hang2025beyondsearch}, treating that reference list as ground truth. This convention has two problems. First, vanilla API search based on keywords misses most of the relevant literature: single and multiple keyword queries to a scholarly index are brittle, and the long-tail structure of citation graphs hides relevant work behind chains of references \citep{kinney2023semanticscholar,priem2022openalex}. Second, the reference list itself is incomplete and idiosyncratic \citep{agarwal2024litllm}, including foundational paper citations and software attribution, in addition to topically relevant papers.

We address both problems and report two findings. (1) A \emph{Deep Research} pipeline that ingests the full query paper and recursively expands its bibliography increases recall by an order of magnitude over vanilla API search. (2) Treating human reference list as a clean ground truth is problematic: a neutral LLM judge \citep{zheng2023judging} rates only 51\% of human citations as moderately relevant or higher, compared with 86--88\% for the strongest automated re-rankers. We probe this gap through the OpenAlex co-authorship graph \citep{priem2022openalex} and find that humans cite direct collaborators 2.5$\times$ more often than the strongest re-rankers.

\begin{figure}[t]
    \centering
    \includegraphics[width=\columnwidth]{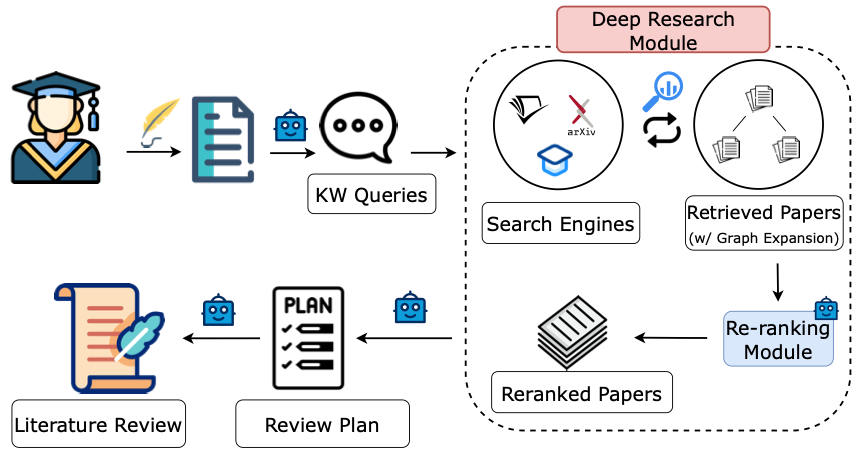}
    \caption{Deep Research pipeline. The system ingests the full query paper, generates LLM-constructed keyword queries, retrieves seed papers from scholarly APIs, expands the candidate set along citation links, and re-ranks the pooled candidates. Full algorithm and prompts in Appendix~\ref{sec:appendix-system}.}
    \label{fig:figure1}
\end{figure}

\paragraph{Contributions.}
(i) A modular Deep Research retrieval pipeline (Figure~\ref{fig:figure1}) for full-text literature search, with order-of-magnitude recall gains over traditional literature search baselines, along with \textsc{RollingEval-Jun25}, a 250-paper benchmark that mitigates data contamination.
(ii) An analysis of human reference lists that documents systematic non-topical citing and provides a framework for re-thinking literature-search ground truth.

\begin{figure*}[t]
    \centering
    \includegraphics[width=0.95\linewidth]{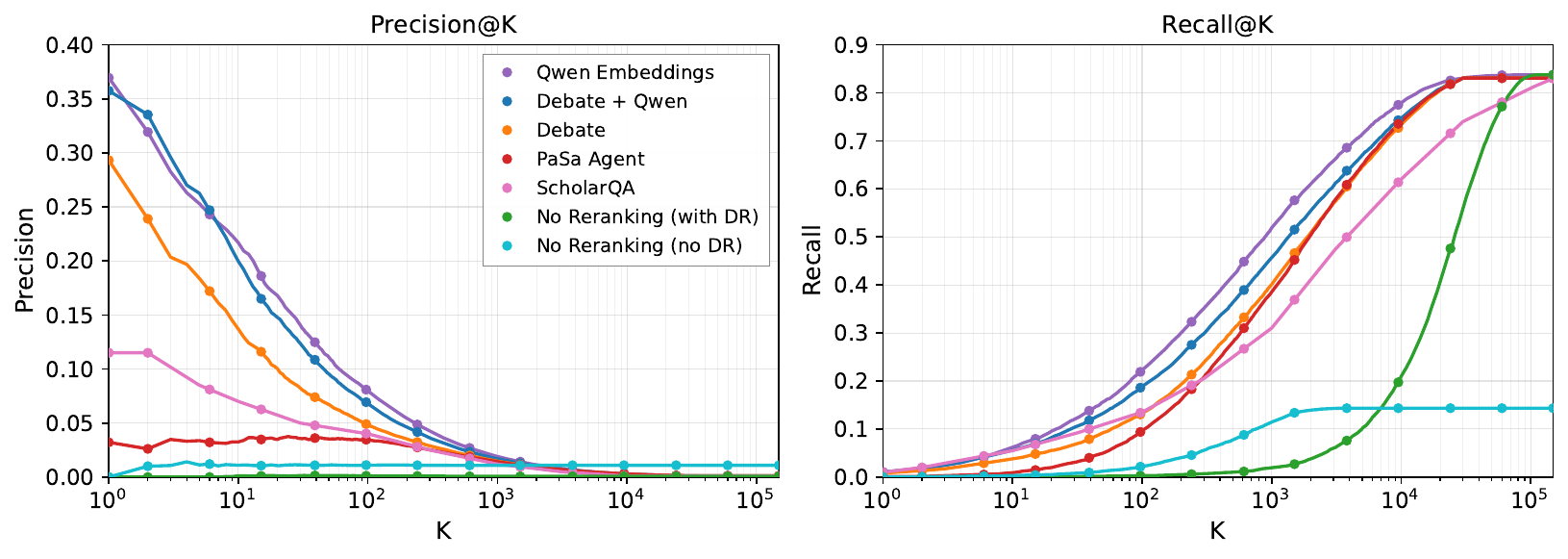}
    \caption{Precision@$K$ (left) and Recall@$K$ (right), $K$ on log scale. Deep Research raises recall by an order of magnitude over normal search; \textsc{Qwen3} embeddings give the strongest top-$K$ precision.}
    \label{fig:pr_curves}
\end{figure*}

\section{Deep Research Pipeline}
\label{sec:method}
The system has two phases: a high-recall retrieval phase and a re-ranking phase. \textbf{Phase~1} prompts an LLM to draft diverse keyword queries for the query document, translates them into provider-specific syntax for arXiv, OpenAlex, and Semantic Scholar \citep{kinney2023semanticscholar,priem2022openalex}, and collects candidates under rate-limit-aware concurrency. \textbf{Phase~2} (Deep Research) expands the candidate set breadth-first along bibliographic references up to a configurable depth and paper budget. One of three interchangeable modules then re-ranks the expanded pool: \textsc{Qwen3} embeddings \citep{qwen3embedding2025}; an LLM-based debate ranking mechanism that produces arguments for and against inclusion before assigning a 0--100 relevance score; or a Debate$+$\textsc{Qwen3} ensemble that combines the two. Algorithm pseudo-code, prompts, and full caching/concurrency details are in Appendix~\ref{sec:appendix-system}.

\section{Setup}
\label{sec:setup}
\textbf{Benchmark.} We construct \textsc{RollingEval-Jun25}, a June~2025 snapshot of the \textsc{RollingEval} series proposed originally by \citep{agarwal2024litllm} to avoid data contamination issues when benchmarking LLMs for literature search. The benchmark comprises 250 computer science arXiv submissions with cleaned full-text and bibliographies. We choose June~2025 as it post-dates the training-data cutoff of every LLM evaluated in this work, mitigating contamination, and the cleaned full-text and bibliography pipeline supports the matched-input head-to-head re-ranker comparison in Section~\ref{sec:results}.
We fetch references through the OpenAlex and SemanticScholar APIs, and fall back on an LLM-based parser that extracts bibliographies directly from PDFs.

\textbf{Metrics.} We report: \textbf{(a)} Precision and Recall against the human reference list, \textbf{(b)} Semantic Relevance (SR), a 0--100 score from an LLM judge \citep{zheng2023judging}, and \textbf{(c)} $\alpha$-nDCG \citep{clarke2008novelty}, the novelty-aware variant of nDCG \citep{jarvelin2002cumulated} that measures how well a ranked list covers diverse topics without redundancy.
Detailed metric definitions are in Appendix~\ref{sec:appendix-metrics}.

The SR judge is \textsc{GPT-OSS-120B} served via vLLM \citep{kwon2023vllm}; for each pair it conditions on the query paper's full text (title, abstract, body) and the candidate's title and abstract, and returns a rubric grade in $\{0,\ldots,5\}$ (see Prompt~\ref{tab:prompt-semrel}), which we then multiply by 20. The re-rankers also score candidates by comparing the query paper's full text with the candidate's title and abstract.

\section{Retrieval Results}
\label{sec:results}
\textbf{Deep Research yields an order-of-magnitude recall gain.} Figure~\ref{fig:pr_curves} plots precision and recall against the human reference list. Vanilla API search (\emph{No Rerank, no DR}) tops out near 15\% recall only at $K{\approx}10^5$ and never exceeds it, while every Deep Research variant exceeds 80\% recall by $K{=}10^4$. \textsc{Qwen3} embeddings achieve the strongest top-$K$ precision (0.37 at $K{=}1$), closely followed by the Debate+Qwen ensemble. We also include \textsc{PaSa} \citep{he2025pasa} and \textsc{ScholarQA} \citep{singh2025ai2scholarqa} as deep-research baselines. We run both on \textsc{RollingEval-Jun25} with the query paper's abstract as input; each system then produces its ranking through its native retrieval and scoring pipeline.
Table~\ref{tab:retrieval} summarizes precision and recall at four representative cutoffs. The best Deep Research variant nearly triples recall at $K{=}100$ over an arXiv-API baseline and reaches 51.9\% recall at $K{=}1000$.

\begin{table}[t]
\centering
\small
\setlength{\tabcolsep}{3pt}
\resizebox{\columnwidth}{!}{
\begin{tabular}{lcccccc}
\toprule
& \multicolumn{3}{c}{\textbf{Precision}} & \multicolumn{3}{c}{\textbf{Recall}} \\
\cmidrule(lr){2-4} \cmidrule(lr){5-7}
\textbf{Method} & @20 & @100 & @1k & @20 & @100 & @1k \\
\midrule
No Rerank (no DR) & 0.1 & 0.1 & 0.0 & 0.0 & 0.3 & 1.9 \\
\textsc{PaSa}       & 5.3 & 4.5 & 1.5 & 1.3 & 7.4 & 27.8 \\
\textsc{ScholarQA} & 5.7 & 4.0 & 1.2 & 7.7 & 13.5 & 31.0 \\
Debate             & 9.6 & 4.5 & 1.5 & 5.1 & 12.2 & 39.9 \\
Debate$+$Qwen3     & 14.7 & 6.6 & 1.7 & 8.1 & 18.1 & 46.2 \\
\textsc{Qwen3} emb.& \textbf{16.8} & \textbf{8.0} & \textbf{1.9} & \textbf{9.6} & \textbf{22.2} & \textbf{51.9} \\
\bottomrule
\end{tabular}}
\caption{Precision and recall at $K \in \{20, 100, 1{,}000\}$. Best column in bold.}
\label{tab:retrieval}
\end{table}

\paragraph{Diversity and ranking quality.}
Beyond precision and recall, we ask whether highly relevant items are concentrated near the top of each ranked list. Table~\ref{tab:diversity} reports $\alpha$-nDCG \citep{clarke2008novelty} and the LLM-judged Semantic Relevance score at $K \in \{10, 100, 1000\}$. Debate (without Qwen) achieves the strongest $\alpha$-nDCG at $K{=}100$ and $K{=}1000$, suggesting that prompt-based scoring leads to higher diversity and relevant items more evenly through the ranked list than embedding similarity. The Debate$+$\textsc{Qwen3} ensemble has the highest top-10 Semantic Relevance (67.9), validating it as the strongest top-of-list retriever, while \textsc{Qwen3} alone maintains the highest Semantic Relevance at $K{=}100$ and $K{=}1000$, indicating better long-tail behavior. The ground-truth row anchors the comparison: the human reference list itself sits at 47--51 on Semantic Relevance across $K$, well below the AI re-rankers in the top-10/100 buckets, a discrepancy we examine next.

\begin{table}[t]
\centering
\small
\setlength{\tabcolsep}{4pt}
\resizebox{\columnwidth}{!}{
\begin{tabular}{lccccccc}
\toprule
& \multicolumn{3}{c}{$\boldsymbol{\alpha}$\textbf{-nDCG}} & \multicolumn{3}{c}{\textbf{Sem.\ Rel.}} \\
\cmidrule(lr){2-4} \cmidrule(lr){5-7}
\textbf{Method} & @10 & @100 & @1k & @10 & @100 & @1k \\
\midrule
No Rerank (no DR)   & 15.3 & 41.2 & 62.1 &  9.2 & 17.3 & 17.4 \\
Debate$+$\textsc{Qwen3}& 53.2 & 76.2 & 87.9 & \textbf{67.9} & 53.8 & 33.2 \\
Debate              & \textbf{54.1} & \textbf{79.7} & \textbf{89.1} & 55.4 & 39.8 & 27.1 \\
\textsc{Qwen3} emb. & 47.6 & 65.3 & 77.8 & 67.1 & \textbf{57.6} & \textbf{47.9} \\
\midrule
Human References         & --- & --- & --- & 47.3 & 50.7 & 50.8 \\
\bottomrule
\end{tabular}}
\caption{$\alpha$-nDCG and Semantic Relevance at $K \in \{10, 100, 1000\}$. Higher is better.}
\label{tab:diversity}
\end{table}

\section{Human References = Ground Truth?}
\label{sec:human-analysis}

Recall against the human reference list is a coherent metric only if that list is itself reliable. We test this by asking a neutral LLM judge to rate the topical relevance of every cited paper, then comparing the resulting distribution against the same judge's ratings of AI re-ranker outputs at matched list length.

\textbf{Setup.} For each of the 250 query papers $P$ with $N_P$ human references, we evaluate all human citations and the top-$N_P$ predictions from each re-ranker. This list-length parity is essential: a longer AI list would trivially include more low-scoring tail items. We aggregate 9,204 human (query, reference) pairs and 36,692 AI pairs.

\begin{figure}[t]
    \centering
    \includegraphics[width=\columnwidth]{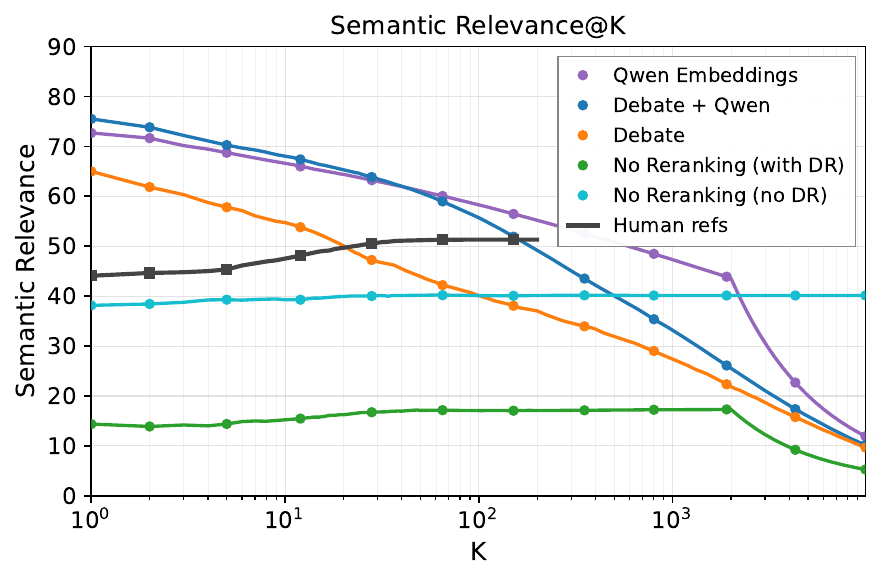}
    \caption{Cumulative semantic relevance over the top-$K$ retrieved candidates, with the SR judge applied to at most $K_{\max}{=}1000$ items per query. Human references saturate near 51 by $K{=}50$; AI re-rankers start substantially higher (73--75 at $K{=}1$) and stay above the human curve across the full $K \leq 1000$ scoring window.}
    \label{fig:sr_curves}
\end{figure}

Table~\ref{tab:topnp_score_dist} shows the matched-$K$ distribution and Figure~\ref{fig:sr_curves} the full curve. Restricted to the same list length, human references score $\geq 60$ in 51.4\% of cases, while \textsc{Qwen3} and Debate+Qwen reach 86.0\% and 87.8\%. The disparity concentrates at score $=40$, which captures 38.2\% of human references but only 10.2--13.7\% of what the top re-rankers retrieve.
On the LLM judge's topical-relevance axis, the strongest re-rankers' top-$N_P$ lists score higher than the human-curated list at matched length; this speaks to one axis of citation quality, not an overall verdict on the human list.
Figure~\ref{fig:sr_curves} makes this concrete across $K$: the AI re-rankers start well above the human saturation level and remain above it throughout the $K \leq 1000$ scoring window, with their cumulative scores gradually diluting toward the matched-$K$ means in Table~\ref{tab:topnp_score_dist} as the top-of-list signal blends with lower-relevance tail items.

\begin{table}[t]
\centering
\small
\setlength{\tabcolsep}{4pt}
\resizebox{\columnwidth}{!}{
\begin{tabular}{lcccc}
\toprule
\textbf{Source} & \textbf{Mean} & \textbf{$\leq 40$} & \textbf{$\geq 60$} & $n$ \\
\midrule
Human                & 51.2 & 48.6\% & 51.4\% & 9{,}204 \\
\textsc{Qwen3} emb.  & 62.4 & 14.0\% & 86.0\% & 9{,}173 \\
Debate+\textsc{Qwen3}& 62.4 & 12.2\% & 87.8\% & 9{,}173 \\
Debate               & 45.6 & 37.9\% & 62.1\% & 9{,}173 \\
No Rerank (no DR)    & 17.3 & 97.7\% & 2.3\%  & 9{,}173 \\
\bottomrule
\end{tabular}}
\caption{Semantic relevance distribution at matched top-$N_P$. $n$ is the total number of (query, reference) pairs.}
\label{tab:topnp_score_dist}
\end{table}

\textbf{Is the gap a citation-collusion artifact?}
A natural hypothesis is that humans cite collaborators. We test this on the OpenAlex co-authorship graph \citep{priem2022openalex}, classifying each (query, candidate) pair by the minimum hop distance between author sets, $d \in \{0, 1, 2, 3, {\geq}4\}$ (formal definition and BFS data requirements in Appendix~\ref{sec:appendix-coauthor}). Table~\ref{tab:coauthor_distance} reports the distribution.

The bias concentrates at short hops and decays with radius. At $d{=}0$ (direct co-authorship) humans are 2.5$\times$ more likely than the strongest re-rankers to cite themselves or a co-author (5.13\% vs.\ 1.9--2.1\%). The cumulative within-one-hop rate is 12.2\% for humans vs.\ 6.9--7.4\% for the best AI methods, a 1.6$\times$ gap. By $d{\leq}2$ the gap narrows to 1.1$\times$ (53.6\% vs.\ 46.6--48.7\%), and by $d{\leq}3$ all sources converge near 80--83\%, consistent with the small-world structure of the academic graph. Conditional mean scores are essentially flat across distance classes for humans (56.1, 52.6, 51.2, 51.2, 50.9 for $d{=}0$ through $d{\geq}4$): the network proximity of a citation does not predict its topical relevance.

\begin{table}[t]
\centering
\small
\setlength{\tabcolsep}{3pt}
\resizebox{\columnwidth}{!}{
\begin{tabular}{lccccc}
\toprule
\textbf{Source} & $d{=}0$ & $d{=}1$ & $d{=}2$ & $d{=}3$ & $d{\geq}4$ \\
\midrule
Human              & 5.13 & 7.02 & 41.43 & 29.13 & 17.28 \\
\textsc{Qwen3}     & 2.05 & 5.36 & 41.25 & 34.14 & 17.19 \\
Debate+\textsc{Qwen3}& 1.92 & 4.98 & 39.69 & 36.06 & 17.35 \\
Debate             & 1.52 & 4.32 & 38.18 & 36.72 & 19.27 \\
No Rerank (no DR)  & 0.09 & 2.65 & 36.87 & 40.17 & 20.22 \\
\bottomrule
\end{tabular}}
\caption{Co-authorship-graph distance at matched top-$N_P$, OpenAlex graph. $d{=}0$: direct co-authorship; higher $d$: shortest co-author path.}
\label{tab:coauthor_distance}
\end{table}

\textbf{Interpretation.} Both effects are individually defensible but together explain the gap. The network bias is concentrated at direct and one-hop collaborators, where humans diverge sharply from the re-rankers (5.13\% vs.\ 1.9--2.1\% at $d{=}0$; 12.2\% vs.\ 6.9--7.4\% within one hop); from $d{\geq}2$ onward the distributions converge, so network proximity drives a clean but bounded share of the disparity. Citing close collaborators is appropriate where the work is genuinely relevant, but at this magnitude it suggests authors should put proportionally more effort into capturing the broader landscape beyond their immediate clique. On the topical axis, foundational background (Transformer, ResNet, BERT \citep{vaswani2017attention,he2016resnet,devlin2019bert}), classical algorithms (Adam \citep{kingma2015adam}), and software or library attribution are all defensible reasons to cite a paper whose abstract carries no topical signal; what is harder to defend is that citations of this kind make up nearly half (48.6\%) of human reference lists at matched length, against 12--14\% for the strongest re-rankers. Eighteen representative examples are in Appendix~\ref{sec:appendix-qualitative}. The practical implication is two-fold: literature-search benchmarks should not treat human reference lists as a clean ground truth, and authors should treat the topical-relevance share of their bibliography as a property worth checking. Future work should also pair the topical rubric with citation-context spans from the query paper, or otherwise model the reasons humans cite.

\section{Discussion and Conclusion}
Our results argue for separating two questions the community usually conflates. \emph{Did the system retrieve the papers the authors cited?} is a coverage question against a known-imperfect target. \emph{Did the system retrieve papers that a neutral reader would judge relevant?} is a different, \emph{complementary} question. The two diverge sharply: Deep Research dominates on the first, while the human-curated list scores lower than the top re-rankers on the second. These two findings are intertwined rather than competing: the order-of-magnitude recall gain matters precisely because it expands the candidate pool from which any retrieval system, including SR-aligned ones, must draw, so a higher-recall pipeline raises the achievable ceiling on whichever metric the community ultimately adopts as ground truth. Neither metric is strictly better: recall against human citations captures dimensions an LLM-judge cannot justify (foundational and tool/library attribution, methodological lineage), while the topical rubric captures dimensions that human reference lists undersample (long-tail topically-related work, non-collaborator authors). We argue for reporting both, alongside the OpenAlex-graph diagnostic, rather than treating either in isolation as ground truth. We will release \textsc{RollingEval-Jun25}, the Deep Research pipeline, the LLM-judge prompts, and the OpenAlex distance analysis as a reproducible scaffold for re-examining literature-search ground truth.

\section*{Limitations}
The LLM judge is a single neutral model; rubric-conditioned scoring may drift across model and prompt versions, and we do not currently model citation context (in-text spans surrounding the citation). Validating against human relevance ratings on a labelled subsample remains a natural next step. The co-authorship graph fetch caps each author at 1{,}000 most recent works, so we under-count distances through hyper-prolific intermediaries, which biases the reported gap in a conservative direction. Bibliography extraction relies on LLM agents and can carry residual noise. The \textsc{RollingEval} benchmark consists only of computer science papers; extending it to biomedical, social-science, and humanities papers would test cross-domain generality. Finally, the analysis covers only arXiv-indexed references; we exclude non-arXiv citations.

\section*{Potential Risks}
The Semantic Relevance judge and Deep Research pipeline we release target legitimate literature discovery, but the same topical-relevance rubric could be misapplied to gate citations during peer review, to filter submissions, or to automate rejection, uses for which a single LLM judge is not a defensible standard. Our results show the rubric and the human reference list disagree systematically along measurable axes; both should serve as complementary diagnostics, not as evaluation thresholds. The OpenAlex co-authorship analysis surfaces network proximity between authors that is computable from public data but that researchers may reasonably consider private; we report only aggregate distributions and avoid disclosing individual-paper proximities. Finally, \textsc{RollingEval-Jun25} comprises arXiv preprints chosen to postdate the training cutoff of every LLM we evaluate; its contamination-resistance guarantee weakens as models update, so the snapshot will need periodic refreshes. Additionally, extending it to include articles from other domains medical, biology, or physics might help study its generalizability.

\bibliography{custom}

@article{
agarwal2024litllm,
title={Lit{LLM}s, {LLM}s for Literature Review: Are we there yet?},
author={Shubham Agarwal and Gaurav Sahu and Abhay Puri and Issam H. Laradji and Krishnamurthy Dj Dvijotham and Jason Stanley and Laurent Charlin and Christopher Pal},
journal={Transactions on Machine Learning Research},
issn={2835-8856},
year={2025},
url={https://openreview.net/forum?id=heeJqQXKg7},
note={}
}

@inproceedings{he2025pasa,
    title = "{P}a{S}a: An {LLM} Agent for Comprehensive Academic Paper Search",
    author = "He, Yichen  and
      Huang, Guanhua  and
      Feng, Peiyuan  and
      Lin, Yuan  and
      Zhang, Yuchen  and
      Li, Hang  and
      E, Weinan",
    editor = "Che, Wanxiang  and
      Nabende, Joyce  and
      Shutova, Ekaterina  and
      Pilehvar, Mohammad Taher",
    booktitle = "Proceedings of ACL",
    month = jul,
    year = "2025",
    address = "Vienna, Austria",
    publisher = "Association for Computational Linguistics",
    url = "https://aclanthology.org/2025.acl-long.572/",
    doi = "10.18653/v1/2025.acl-long.572",
    pages = "11663--11679",
    ISBN = "979-8-89176-251-0",
}

@inproceedings{singh2025ai2scholarqa,
    title = "{Ai2 ScholarQA}: Organized Literature Synthesis with Attribution",
    author = "Singh, Amanpreet  and
      Chang, Joseph Chee  and
      Haddad, Dany  and
      Naik, Aakanksha  and
      Hwang, Jena D.  and
      Kinney, Rodney  and
      Weld, Daniel S  and
      Downey, Doug  and
      Feldman, Sergey",
    editor = "Mishra, Pushkar  and
      Muresan, Smaranda  and
      Yu, Tao",
    booktitle = "Proceedings of ACL",
    month = jul,
    year = "2025",
    address = "Vienna, Austria",
    publisher = "Association for Computational Linguistics",
    url = "https://aclanthology.org/2025.acl-demo.49/",
    doi = "10.18653/v1/2025.acl-demo.49",
    pages = "513--523",
    ISBN = "979-8-89176-253-4",
}

@inproceedings{ajith2024litsearch,
  title={Litsearch: A retrieval benchmark for scientific literature search},
  author={Ajith, Anirudh and Xia, Mengzhou and Chevalier, Alexis and Goyal, Tanya and Chen, Danqi and Gao, Tianyu},
  booktitle={Proceedings of the 2024 Conference on EMNLP},
  pages={15068--15083},
  year={2024}
}

@inproceedings{cohan2020specter,
    title = "{SPECTER}: Document-level Representation Learning using Citation-informed Transformers",
    author = "Cohan, Arman  and
      Feldman, Sergey  and
      Beltagy, Iz  and
      Downey, Doug  and
      Weld, Daniel",
    editor = "Jurafsky, Dan  and
      Chai, Joyce  and
      Schluter, Natalie  and
      Tetreault, Joel",
    booktitle = "Proceedings of the 58th Annual Meeting of the ACL",
    month = jul,
    year = "2020",
    address = "Online",
    publisher = "Association for Computational Linguistics",
    url = "https://aclanthology.org/2020.acl-main.207/",
    doi = "10.18653/v1/2020.acl-main.207",
    pages = "2270--2282",
}

@article{qwen3embedding2025,
  title={Qwen3 embedding: Advancing text embedding and reranking through foundation models},
  author={Zhang, Yanzhao and Li, Mingxin and Long, Dingkun and Zhang, Xin and Lin, Huan and Yang, Baosong and Xie, Pengjun and Yang, An and Liu, Dayiheng and Lin, Junyang and others},
  journal={arXiv preprint arXiv:2506.05176},
  year={2025}
}

@article{kinney2023semanticscholar,
  title={The semantic scholar open data platform},
  author={Kinney, Rodney and Anastasiades, Chloe and Authur, Russell and Beltagy, Iz and Bragg, Jonathan and Buraczynski, Alexandra and Cachola, Isabel and Candra, Stefan and Chandrasekhar, Yoganand and Cohan, Arman and others},
  journal={arXiv preprint arXiv:2301.10140},
  year={2023}
}

@article{priem2022openalex,
  title={OpenAlex: A fully-open index of scholarly works, authors, venues, institutions, and concepts},
  author={Priem, Jason and Piwowar, Heather and Orr, Richard},
  journal={arXiv preprint arXiv:2205.01833},
  year={2022}
}

@inproceedings{clarke2008novelty,
  title={Novelty and diversity in information retrieval evaluation},
  author={Clarke, Charles LA and Kolla, Maheedhar and Cormack, Gordon V and Vechtomova, Olga and Ashkan, Azin and B{\"u}ttcher, Stefan and MacKinnon, Ian},
  booktitle={Proceedings of the 31st annual international ACM SIGIR conference on Research and development in information retrieval},
  pages={659--666},
  year={2008}
}

@article{jarvelin2002cumulated,
  title={Cumulated gain-based evaluation of IR techniques},
  author={J{\"a}rvelin, Kalervo and Kek{\"a}l{\"a}inen, Jaana},
  journal={ACM Transactions on Information Systems (TOIS)},
  volume={20},
  number={4},
  pages={422--446},
  year={2002},
  publisher={ACM New York, NY, USA}
}

@article{mcinnes2017hdbscan,
  title={hdbscan: Hierarchical density based clustering.},
  author={McInnes, Leland and Healy, John and Astels, Steve and others},
  journal={Journal of Open Source Software},
  volume={2},
  number={11},
  pages={205},
  year={2017}
}

@inproceedings{kwon2023vllm,
  title={Efficient memory management for large language model serving with pagedattention},
  author={Kwon, Woosuk and Li, Zhuohan and Zhuang, Siyuan and Sheng, Ying and Zheng, Lianmin and Yu, Cody Hao and Gonzalez, Joseph and Zhang, Hao and Stoica, Ion},
  booktitle={Proceedings of SOSP},
  pages={611--626},
  year={2023}
}

@inproceedings{zheng2023judging,
  title     = {Judging {LLM}-as-a-Judge with {MT-Bench} and Chatbot Arena},
  author    = {Zheng, Lianmin and Chiang, Wei-Lin and Sheng, Ying and Zhuang, Siyuan and Wu, Zhanghao and Zhuang, Yonghao and Lin, Zi and Li, Zhuohan and Li, Dacheng and Xing, Eric P. and Zhang, Hao and Gonzalez, Joseph E. and Stoica, Ion},
  booktitle = {NeurIPS Datasets and Benchmarks},
  year      = {2023}
}

@inproceedings{devlin2019bert,
  title={Bert: Pre-training of deep bidirectional transformers for language understanding},
  author={Devlin, Jacob and Chang, Ming-Wei and Lee, Kenton and Toutanova, Kristina},
  booktitle={Proceedings of the 2019 conference of the NAACL: HLT},
  pages={4171--4186},
  year={2019}
}

@inproceedings{kingma2015adam,
  title     = {{Adam}: A Method for Stochastic Optimization},
  author    = {Kingma, Diederik P. and Ba, Jimmy},
  booktitle = {ICLR},
  year      = {2015}
}

@article{vaswani2017attention,
  title={Attention is all you need},
  author={Vaswani, Ashish and Shazeer, Noam and Parmar, Niki and Uszkoreit, Jakob and Jones, Llion and Gomez, Aidan N and Kaiser, {\L}ukasz and Polosukhin, Illia},
  journal={Advances in NeurIPS},
  volume={30},
  year={2017}
}

@inproceedings{he2016resnet,
  title={Deep residual learning for image recognition},
  author={He, Kaiming and Zhang, Xiangyu and Ren, Shaoqing and Sun, Jian},
  booktitle={Proceedings of CVPR},
  pages={770--778},
  year={2016}
}

@article{robertson2009bm25,
  title   = {The Probabilistic Relevance Framework: {BM25} and Beyond},
  author  = {Robertson, Stephen and Zaragoza, Hugo},
  journal = {Foundations and Trends in Information Retrieval},
  volume  = {3}, number = {4}, pages = {333--389},
  year    = {2009}
}

@inproceedings{karpukhin2020dpr,
    title = "Dense Passage Retrieval for Open-Domain Question Answering",
    author = "Karpukhin, Vladimir  and
      Oguz, Barlas  and
      Min, Sewon  and
      Lewis, Patrick  and
      Wu, Ledell  and
      Edunov, Sergey  and
      Chen, Danqi  and
      Yih, Wen-tau",
    editor = "Webber, Bonnie  and
      Cohn, Trevor  and
      He, Yulan  and
      Liu, Yang",
    booktitle = "Proceedings of EMNLP",
    month = nov,
    year = "2020",
    address = "Online",
    publisher = "Association for Computational Linguistics",
    url = "https://aclanthology.org/2020.emnlp-main.550/",
    doi = "10.18653/v1/2020.emnlp-main.550",
    pages = "6769--6781"
}

@article{lala2023paperqa,
  title={{PaperQA}: Retrieval-augmented generative agent for scientific research},
  author={L{\'a}la, Jakub and O'Donoghue, Odhran and Shtedritski, Aleksandar and Cox, Sam and Rodriques, Samuel G and White, Andrew D},
  journal={arXiv preprint arXiv:2312.07559},
  year={2023}
}

@inproceedings{hang2025beyondsearch,
  title     = {Beyond Search: Measuring {LLM} Performance for Scientific Literature Discovery},
  author    = {Hang, Ching Nam and Yu, Pei-Duo and Tan, C. and Chiu, D.},
  booktitle = {IEEE TALE},
  year      = {2025},
  doi       = {10.1109/TALE66047.2025.11346619}
}

@article{lei2025rhinoinsight,
  title   = {{RhinoInsight}: Improving Deep Research through Control Mechanisms for Model Behavior and Context},
  author  = {Lei, Yu and Si, Shuzheng and Wang, Wei and Wu, Yifei and Chen, Gang and Qi, Fanchao and Sun, Maosong},
  journal = {arXiv preprint arXiv:2511.18743},
  year    = {2025}
}

@article{zhang2025semrank,
  title   = {Scientific Paper Retrieval with {LLM}-Guided Semantic-Based Ranking},
  author  = {Zhang, Yunyi and Yang, Ruozhen and Jiao, Siqi and Kang, SeongKu and Han, Jiawei},
  journal = {arXiv preprint arXiv:2505.21815},
  year    = {2025}
}

@inproceedings{zhang2026toolagent,
  title     = {Tool-Augmented Multi-Turn Academic Paper Recommendation via Reinforcement Learning},
  author    = {Zhang, Jiarong and Du, Junping and Xue, Zhe and Ye, Guanhua and Shao, Yingxia},
  booktitle = {IEEE BigComp},
  year      = {2026},
  doi       = {10.1109/BigComp68355.2026.00031}
}

\appendix

\section{Background and Related Work}
\label{sec:appendix-background}

\paragraph{Scholarly APIs and index-based retrieval.}
Scholarly APIs such as OpenAlex \citep{priem2022openalex}, Semantic Scholar \citep{kinney2023semanticscholar}, and arXiv remain the backbone of scalable literature discovery, but require careful query design to maximize recall under strict rate limits, and a single API call against them typically saturates well below the topical neighborhood of the query paper. We therefore use these APIs only as the entry-point fetch layer, drafting an LLM-constructed keyword set against them in Phase~1 and then expanding breadth-first along bibliographic references in Phase~2 (Section~\ref{sec:method}).

\paragraph{Dense retrievers and re-rankers.}
Dense retrievers such as DPR \citep{karpukhin2020dpr} and BM25 \citep{robertson2009bm25}, together with citation-aware embeddings like SPECTER \citep{cohan2020specter}, provide stronger semantic matches than keyword search but trade off efficiency at scale; more recent LLM-guided schemes such as SemRank \citep{zhang2025semrank} graft concept-level matching on top of a dense retriever. We treat the re-ranker as an interchangeable module: \textsc{Qwen3-Embedding-8B} \citep{qwen3embedding2025}, an LLM-based Debate ranker, and their ensemble plug into the same final stage of the pipeline and are compared head-to-head under matched inputs and matched candidate pools.

\paragraph{LLM-based retrieval agents.}
LLM-based retrieval agents such as PaSa \citep{he2025pasa}, Ai2 ScholarQA \citep{singh2025ai2scholarqa}, PaperQA \citep{lala2023paperqa}, RhinoInsight \citep{lei2025rhinoinsight}, and earlier LitLLM \citep{agarwal2024litllm} combine API search with LLM scoring, control mechanisms, or bibliography expansion, and \citet{zhang2026toolagent} extend the family to a multi-turn, tool-using policy trained with reinforcement learning. We hold the orchestration policy constant (keyword expansion followed by bibliography-graph expansion) and instead vary (i) full-text input vs.\ abstract-only input and (ii) the choice of re-ranker, which isolates the contribution of full-paper context and citation-graph reach from the choice of agent policy.

\paragraph{Evaluating LLM literature search.}
On the evaluation side, the \textsc{LitSearch} benchmark \citep{ajith2024litsearch} and the three-metric framework of \citet{hang2025beyondsearch}, which scores GPT-4, Gemini~2.5, and DeepSeek-V3 in vanilla and ``deep research'' configurations, both anchor quality on recall against the human reference list. Our analysis treats that recall as one diagnostic axis, pairs it with a semantic-relevance rubric scored by a neutral LLM judge, and adds an OpenAlex co-authorship-graph distance metric (Section~\ref{sec:human-analysis}); the first two metrics disagree systematically on real data, and the co-authorship analysis attributes a measurable portion of the gap to network bias in human citing rather than to retrieval quality alone.

\section{System Details}
\label{sec:appendix-system}
\subsection{Phase 1: Retrieval with LLM-Constructed Queries}
Given a query document, we prompt an LLM to draft a diverse keyword query set, then normalize each query into the provider-specific syntax for arXiv, OpenAlex, and Semantic Scholar \citep{kinney2023semanticscholar,priem2022openalex}. We pre-process inputs to remove inline citations and bibliography content before prompting, and cache all generated queries by input- and model-specific hash.

\subsection{Phase 2: Citation Graph Expansion}
We expand the seed set breadth-first along bibliographic references up to a configurable depth $D_{max}$ (=3 in our experiments) and paper budget $N_{max}$. When references from APIs like SemanticScholar and OpenAlex are incomplete, we fall back on LLM-based PDF bibliography parsing, which first extracts the bibliography section from the paper and resolves each entry through its title and authors using SemanticScholar and OpenAlex. Algorithm~\ref{alg:deep_research} states the procedure.

\begin{algorithm}
\caption{Deep Research: Citation Graph Expansion}
\label{alg:deep_research}
\begin{algorithmic}[1]
\REQUIRE Initial Seed Papers $P_{seed}$, Max Depth $D_{max}$, Max Papers $N_{max}$
\STATE $Q \leftarrow \text{new Queue}()$
\STATE $P_{final} \leftarrow P_{seed}$
\STATE $V_{ids} \leftarrow \{\text{id}(p) \text{ for } p \in P_{seed}\}$
\STATE \textbf{for each} $p \in P_{seed}$ \textbf{do}
    \STATE $Q.\text{enqueue}((p, 0))$
\STATE \textbf{end for}
\WHILE{$Q$ is not empty \textbf{and} $|P_{final}| < N_{max}$}
    \STATE $(p_{curr}, d_{curr}) \leftarrow Q.\text{dequeue}()$
    \IF{$d_{curr} \geq D_{max}$}
        \STATE \textbf{continue}
    \ENDIF
    \STATE $P_{cand} \leftarrow \text{GetReferences}(p_{curr})$

    \FOR{\textbf{each} $p_{new} \in P_{cand}$}
        \IF{$\text{id}(p_{new}) \notin V_{ids}$}
            \STATE $V_{ids}.\text{add}(\text{id}(p_{new}))$
            \STATE $P_{final}.\text{add}(p_{new})$
            \STATE $Q.\text{enqueue}((p_{new}, d_{curr} + 1))$
            \IF{$|P_{final}| \geq N_{max}$}
                \STATE \textbf{break}
            \ENDIF
        \ENDIF
    \ENDFOR
\ENDWHILE
\RETURN $P_{final}$
\end{algorithmic}
\end{algorithm}

\subsection{Re-ranking Modules}
\begin{itemize}
\item \textsc{Qwen3} embeddings \citep{qwen3embedding2025} served via vLLM \citep{kwon2023vllm}, scoring each candidate by cosine similarity between the instruction-conditioned query embedding and the bare candidate embedding.
\item LLM debate ranking: an LLM agent scores each candidate (0--100) with arguments-for/against conditioned on the query paper's full text. Batches run asynchronously under a concurrency semaphore.
\item Debate$+$\textsc{Qwen3} ensemble: averages the two normalized scores into a single per-candidate ranking.
\end{itemize}

\subsection{Engineering Notes}
All experiments use asynchronous execution with bounded parallelism, query/result caching keyed by input hash, and YAML-driven configuration. API semaphores respect provider limits while maintaining throughput.

\section{Prompt Templates}
\label{sec:appendix-prompts}

This appendix lists the four prompts the Deep Research pipeline and LLM-as-a-judge evaluation use: (i) keyword generation for Phase~1 retrieval (Prompt~\ref{tab:prompt-keyword}), (ii) LLM debate ranking (Prompt~\ref{tab:prompt-debate}), (iii) the instruction prefix for \textsc{Qwen3} embedding queries (Prompt~\ref{tab:prompt-qwen3}), and (iv) the semantic-relevance judge (Prompt~\ref{tab:prompt-semrel}) that produces all scores in Sections~\ref{sec:results}--\ref{sec:human-analysis}.

\subsection{Keyword Generation}
\label{sec:appendix-prompt-keyword}
Phase~1 prompts an LLM to draft a diverse set of search queries given the query paper's abstract.

\begin{table*}[t]
\centering
\small
\renewcommand{\arraystretch}{1.15}
\begin{tabular}{@{}p{0.06\linewidth}p{0.89\linewidth}@{}}
\toprule
\textbf{Role} & \textbf{Content} \\
\midrule
System &
You are a helpful research assistant who is helping with literature review of a research idea.
\\
\midrule
User &
You are a helpful research assistant assisting with a literature review for a research idea. You will be given the abstract of a scientific paper. Your goal is to generate a diverse set of mutually exclusive search queries to help find relevant and citable academic papers using scholarly search engines.

\smallskip
Here is the abstract: \texttt{\{paper\_text\}}

\smallskip
\textbf{Instructions.}
\begin{itemize}\setlength{\itemsep}{1pt}\setlength{\parskip}{0pt}\setlength{\topsep}{2pt}
  \item Generate 10 search queries written in the natural, concise style typically used by researchers using academic search engines (e.g., OpenAlex, Semantic Scholar, or Google Scholar).
  \item Each query should reflect a \textbf{different angle} of the abstract (e.g., method, task, dataset, domain, application, or novelty).
  \item Use a \textbf{variety of keywords and phrasings} to maximize diversity and avoid redundancy.
  \item Focus on \textbf{maximizing recall} (retrieving a broad but relevant set of papers), not just precision.
  \item Avoid stopwords or overly verbose phrasing; use terms that researchers would actually search for.
  \item Return a JSON object with the following structure:
\end{itemize}

\smallskip
\hspace*{1em}{\ttfamily\small\{\newline
\hspace*{2em}"queries": [\newline
\hspace*{3em}"first query here",\newline
\hspace*{3em}"second query here"\newline
\hspace*{2em}]\newline
\hspace*{1em}\}}
\\
\bottomrule
\end{tabular}
\caption{Keyword generation prompt (Phase~1 of Deep Research).}
\label{tab:prompt-keyword}
\end{table*}

\subsection{Debate Ranking}
\label{sec:appendix-prompt-debate}
The debate re-ranker (Table~\ref{tab:prompt-debate}) scores each candidate on a 0--100 scale by first writing arguments for and against citing it. We batch candidates per LLM call and share the system prompt with the keyword-generation step.

\begin{table*}[t]
\centering
\small
\renewcommand{\arraystretch}{1.15}
\begin{tabular}{@{}p{0.06\linewidth}p{0.89\linewidth}@{}}
\toprule
\textbf{Role} & \textbf{Content} \\
\midrule
System &
You are a helpful research assistant who is helping with literature review of a research idea.
\\
\midrule
User &
You are a helpful research assistant. Your task is to rank some papers based on their relevance to a query paper.

\smallskip
Given the query paper:\newline
\hspace*{1em}{\ttfamily\small<query\_paper>\newline
\hspace*{1em}\{query\_full\}\newline
</query\_paper>}

\smallskip
And the following candidate reference paper abstracts:\newline
\hspace*{1em}{\ttfamily\small<candidate\_paper\_abstracts>\newline
\hspace*{1em}\{reference\_papers\}\newline
</candidate\_paper\_abstracts>}

\smallskip
\textbf{Instructions.}
\begin{itemize}\setlength{\itemsep}{1pt}\setlength{\parskip}{0pt}\setlength{\topsep}{2pt}
  \item For EVERY candidate paper, provide a relevance score between 0 and 100 representing the probability that the query paper would cite it.
  \item The relevance score MUST be an integer between 0 and 100 (inclusive).
  \item The score MUST be written using digits only (e.g., 0, 17, 42, 50, 100).
  \item Do NOT write numbers in words (e.g., ``thirty-seven'' or ``fifty'').
  \item If a candidate paper happens to be a duplicate of the query paper, it should receive a score of 0.
  \item Provide arguments for and against citing the candidate paper, extracting supporting sentences from the candidate's abstract.
  \item Format your response for EACH paper using the specified tags below.
\end{itemize}

\smallskip
\textbf{Response format for each paper:}\newline
\hspace*{1em}{\ttfamily\small<arguments\_for>\newline
\hspace*{1em}[paper's id]: [Arguments for including the paper]\newline
\hspace*{1em}Extracted Sentences: "Sentence 1", "Sentence 2", \ldots\newline
</arguments\_for>\newline
<arguments\_against>\newline
\hspace*{1em}[paper's id]: [Arguments for not including the paper]\newline
\hspace*{1em}Extracted Sentences: "Sentence 1", "Sentence 2", \ldots\newline
</arguments\_against>\newline
<probability>\newline
\hspace*{1em}paper\_id: [paper's id]\newline
\hspace*{1em}score: [Final Probability Score Based on the Arguments]/100\newline
</probability>}

\smallskip
Note: your response MUST contain arguments and probabilities for ALL the candidate paper abstracts.
\\
\bottomrule
\end{tabular}
\caption{Debate ranking prompt. Arguments-for/against and a 0--100 score are produced per candidate.}
\label{tab:prompt-debate}
\end{table*}

\subsection{\textsc{Qwen3} Embedding Instruction}
\label{sec:appendix-prompt-qwen3}
Following the recommended usage of \textsc{Qwen3-Embedding-8B} \citep{qwen3embedding2025}, we prefix each query paper with a one-sentence task instruction before embedding, and embed candidate abstracts without one. Cosine similarity between instruction-conditioned query embeddings and bare candidate embeddings produces the final ranking.

\begin{table}[t]
\centering
\small
\renewcommand{\arraystretch}{1.15}
\begin{tabular}{@{}p{0.18\columnwidth}p{0.74\columnwidth}@{}}
\toprule
\textbf{Side} & \textbf{Content} \\
\midrule
Query prefix &
{\ttfamily\small Instruct:} You are a helpful research assistant. Your task is to retrieve the candidates that are highly likely to be cited by the query paper below.

\smallskip
{\ttfamily\small Query:}\newline
\hspace*{1em}\texttt{\{query\_paper\}}
\\
\midrule
Candidate &
{\itshape (no instruction prefix)}
\\
\bottomrule
\end{tabular}
\caption{\textsc{Qwen3-Embedding-8B} instruction prefix applied to the query side only.}
\label{tab:prompt-qwen3}
\end{table}

\subsection{Semantic Relevance Judge}
\label{sec:appendix-prompt-semrel}
Every Semantic Relevance score we report (the matched-$K$ distribution in Table~\ref{tab:topnp_score_dist}, the cumulative curves in Figure~\ref{fig:sr_curves}, and the per-method scores in Table~\ref{tab:diversity}) comes from the prompt in Table~\ref{tab:prompt-semrel}. We show the judge the query paper (title, abstract, optionally full text) and a single candidate (title, abstract); it returns an integer in $\{0, 1, 2, 3, 4, 5\}$, which we multiply by~20 to obtain the 0--100 scale used in the main text.

\begin{table*}[t]
\centering
\small
\renewcommand{\arraystretch}{1.15}
\begin{tabular}{@{}p{0.06\linewidth}p{0.89\linewidth}@{}}
\toprule
\textbf{Role} & \textbf{Content} \\
\midrule
System &
You are an expert academic research assistant. You will be shown a query paper and a candidate paper and your task is to analyze the semantic relevance of the query paper and the candidate paper. Assess how relevant the candidate paper is to the subject matter, research scope, and focus of the query paper. Consider topical overlap, methodological similarity, shared objectives, and whether the candidate contributes meaningfully to the themes of the query paper.
\\
\midrule
User &
You are an expert academic research assistant.

\smallskip
\textbf{Input.}\newline
Query Paper Details:\newline
\hspace*{1em}Title: \texttt{\{query\_title\}}\newline
\hspace*{1em}Abstract: \texttt{\{query\_abstract\}}\newline
\hspace*{1em}Full Paper: \texttt{\{query\_full\}}

\smallskip
Candidate Paper:\newline
\hspace*{1em}Title: \texttt{\{candidate\_title\}}\newline
\hspace*{1em}Abstract: \texttt{\{candidate\_abstract\}}

\smallskip
\textbf{Instructions.} Analyze the semantic relevance of the query paper and the candidate paper. Assess how relevant the candidate paper is to the subject matter, research scope, and focus of the query paper.

\smallskip
\textbf{Task rubric.}
\begin{itemize}\setlength{\itemsep}{1pt}\setlength{\parskip}{0pt}\setlength{\topsep}{2pt}
  \item \textbf{5 (Direct Correspondence):} Candidate directly addresses the same research problem as the query paper.
  \item \textbf{4 (Primary Topical Focus):} Candidate's central theme is closely related to the query paper.
  \item \textbf{3 (Substantial Topical Coverage):} Candidate covers significant aspects of the query paper's domain.
  \item \textbf{2 (Peripheral Topical Treatment):} Candidate addresses the query paper's subject as a secondary element.
  \item \textbf{1 (Tangential Relevance):} Minimal substantive overlap.
  \item \textbf{0 (No Substantive Relevance):} Candidate is from a different domain or research area.
\end{itemize}

\smallskip
\textbf{Output format.}\newline
\hspace*{1em}{\ttfamily\small\{\newline
\hspace*{1em}\ "paper\_to\_paper\_relevance": \{\newline
\hspace*{2em}\ "relevanceScore": 0,\newline
\hspace*{2em}\ "confidenceLevel": 0,\newline
\hspace*{2em}\ "summaryStatement": "\ldots"\newline
\hspace*{1em}\ \}\newline
\}}
\\
\bottomrule
\end{tabular}
\caption{Semantic relevance judge prompt. Returns an integer 0--5 per (query, candidate) pair; we multiply by 20 to obtain the 0--100 scale used throughout the paper.}
\label{tab:prompt-semrel}
\end{table*}

\section{Metrics: Definitions}
\label{sec:appendix-metrics}

\paragraph{Precision and Recall.}
For retrieved set $R@K$ and ground-truth set $G$,
\[
\text{Precision@K} = \frac{|R@K \cap G|}{|R@K|}
\]
\[
\text{Recall@K} = \frac{|R@K \cap G|}{|G|}
\]
We report curves for $K$ up to $1.5{\times}10^5$ to assess long-tail retrieval.

\paragraph{LLM-as-a-judge relevance.}
For each query--candidate pair, an LLM \citep{zheng2023judging} assigns a graded relevance score on a 5-point rubric, which we multiply by 20 to obtain a 0--100 scale. The judge conditions on the query paper and the candidate's title and abstract.

\paragraph{$\boldsymbol{\alpha}$-nDCG.}
We extend nDCG \citep{jarvelin2002cumulated} with the novelty-aware variant of \citet{clarke2008novelty}. For each retrieved document $d$ that belongs to a ground-truth cluster $c$, the contribution at rank $r$ is
\[
\text{gain}(d, r) = \frac{1}{\log_2(r+1)} \cdot (1-\alpha)\,\alpha^{n_c},
\]
where $n_c$ is the number of earlier documents from cluster $c$. We normalize by the ideal gain to obtain $\alpha$-nDCG@K. Clusters come from HDBSCAN \citep{mcinnes2017hdbscan} over Qwen3 embeddings \citep{qwen3embedding2025} of candidate-paper abstracts, with the default \texttt{min\_cluster\_size}, so cluster identity reflects topical neighborhoods of the candidate pool.

\section{Co-Authorship Graph Distance: Definition and Computation}
\label{sec:appendix-coauthor}

This appendix formalizes the co-authorship-graph distance metric from Section~\ref{sec:human-analysis} and documents the data we fetch to compute it.

\paragraph{Graph and distance.} Let $G = (V, E)$ be the undirected co-authorship graph in which $V$ is the set of academic authors and an edge $(a, b) \in E$ exists whenever $a$ and $b$ have co-authored at least one publication. For a query paper $Q$ with author set $A_Q$ and a candidate paper $C$ with author set $A_C$, the pair distance is
\begin{equation}
d(Q, C) \;=\; \min_{a \in A_Q,\; b \in A_C} d_G(a, b),
\end{equation}
where $d_G(\cdot, \cdot)$ is the standard shortest-path distance on $G$, with $d_G(a, a) = 0$ and $d_G(a, b) = \infty$ for disconnected pairs.

\paragraph{Hop neighborhoods.} For a set $S \subseteq V$,
\begin{align*}
L_0(S) &= S, \\
L_{k+1}(S) &= \bigcup_{u \in L_k(S)} N(u),
\end{align*}
where $N(u) = \{v : (u, v) \in E\}$ is the set of co-authors of $u$.

\paragraph{Distance test.} We determine the pair distance by checking each level in increasing order:
\begin{align*}
d{=}0:\;\; & A_Q \cap A_C \neq \emptyset, \\
d{=}1:\;\; & L_1(A_Q) \cap A_C \neq \emptyset, \\
d{=}2:\;\; & \bigl(L_1(A_Q) \cap L_1(A_C) \neq \emptyset\bigr) \lor \\
& \bigl(L_2(A_Q) \cap A_C \neq \emptyset\bigr), \\
d{=}3:\;\; & L_2(A_Q) \cap L_1(A_C) \neq \emptyset.
\end{align*}
We record pairs failing all four tests as $d \geq 4$. The $d{=}2$ disjunction is necessary because a length-2 path takes either form: a shared collaborator, or a second-hop reach.

\paragraph{Data requirements.} We materialize three subsets:
\begin{itemize}
\item $L_1(A_Q)$ requires the co-author list of each of the 1{,}113 unique query authors.
\item $L_1(A_C)$ requires the co-author list of each candidate author that is not already in $L_1(A_Q) \cup A_Q$ (52{,}868 authors after exclusion).
\item $L_2(A_Q)$ requires the co-author list of each author in $L_1(A_Q)$ (114{,}669 authors), which is the dominant cost.
\end{itemize}
$d{=}3$ detection requires \emph{both} the candidate-side fetch and the second-hop expansion: the $d{=}3$ intersection $L_2(A_Q) \cap L_1(A_C)$ draws one factor from each side.

\paragraph{Bounded-degree approximation.} We construct each author's co-author list by aggregating across the OpenAlex works in 5 pages of 200 results. Under-counting can only inflate the estimated distance, so the reported gap is a conservative lower bound.

\paragraph{Identifier reconciliation.} We resolve author identities through OpenAlex identifiers \citep{priem2022openalex}, and map arXiv papers to OpenAlex Works via the batched DOI filter using the prefix \texttt{10.48550/arxiv.}. Of 22{,}935 unique arXiv identifiers, 19{,}990 (87.2\%) resolve; we exclude pairs in which either side fails.

\paragraph{Cost.} OpenAlex's batched author-id filter accepts up to 50 IDs per request at \$0.0001 per request under the prepaid plan. The complete pipeline costs approximately \$1.33 in API spend and 4--5 hours of wall-clock time with parallel fetches.

\section{Qualitative Examples of Low-Scoring Human References}
\label{sec:appendix-qualitative}

Table~\ref{tab:qualitative_examples_appendix} lists eighteen representative human-curated references that receive a low semantic relevance score (0 or 20 on the 0--100 rubric). Examples are sampled to cover the recurring categories of non-topical citation we observe across the 250-paper set: foundational architectures (Inception, ResNet \citep{he2016resnet}, Llama-3, DDPM), classical algorithms (Adam \citep{kingma2015adam}, CMA-ES, GELU, the Concrete distribution), software libraries (PyTorch, scikit-learn, CatBoost), benchmarks (ImageNet, DQN/Atari), and cross-domain references (autonomous-vehicle localization, asset pricing, OpenAlex). These citations are common and typically well-motivated in the body, yet their abstracts contain no topical signal that an abstract-conditioned judge can use.

\begin{table*}[t]
\centering
\footnotesize
\setlength{\tabcolsep}{4pt}
\begin{tabular}{p{2.4cm} p{2.7cm} p{3.0cm} p{6.3cm} c}
\toprule
\textbf{Cited work} & \textbf{Cited topic} & \textbf{Citing paper's focus} & \textbf{Judge rationale (abridged)} & \textbf{Score} \\
\midrule
\addlinespace
CMA-ES tutorial\newline (1604.00772) & Continuous black-box optimization & Unsupervised environment design for RL & ``Continuous black-box optimization; does not address unsupervised environment design or curriculum learning for RL.'' & 0 \\
\addlinespace
Inception\newline (1512.00567) & Convolutional vision architecture & Multi-perspective disagreement-aware NLP modeling & ``Computer vision architecture improvements; no overlap with multi-perspective, disagreement-aware NLP modeling.'' & 0 \\
\addlinespace
Concrete distribution\newline (1611.00712) & Continuous relaxation of discrete random variables & Heterogeneous graph contrastive learning & ``Continuous relaxations of discrete random variables; unrelated to heterogeneous graph contrastive learning.'' & 0 \\
\addlinespace
DDPM\newline (2006.11239) & Diffusion probabilistic image synthesis & Volumetric neural body modeling and collision handling & ``Diffusion probabilistic image synthesis; unrelated to volumetric neural body modeling, contacts, and collision handling.'' & 0 \\
\addlinespace
No-norm Transformers\newline (2503.10622) & Eliminating normalization in Transformers & Test-time adaptation for time-series forecasting & ``Eliminating normalization in Transformers; unrelated to test-time adaptation or time-series forecasting.'' & 0 \\
\addlinespace
OpenAlex\newline (2301.10140) & Open scholarly data platform and knowledge graph & Multi-agent LLMs for language model architecture discovery & ``Open scholarly data platform and knowledge graph; unrelated to multi-agent LLM architecture discovery.'' & 0 \\
\addlinespace
Asset-pricing ML\newline (2403.06779) & Machine learning in asset pricing (finance) & Multi-agent LLMs for language model architecture discovery & ``Machine learning applications in asset pricing (finance); unrelated to autonomous LM architecture discovery.'' & 0 \\
\addlinespace
AV localization\newline (1906.01061) & Autonomous-vehicle localization requirements & LLM-powered multimodal accessibility agents & ``Autonomous vehicle localization requirements; unrelated to LLM-powered multimodal accessibility agents.'' & 0 \\
\addlinespace
Llama-3\newline (2407.21783) & Foundation language model and capabilities & Attention-logit analysis and contextual compression & ``Describes a new LLM and its capabilities; lacks focus on attention-logit analysis or compression/attribution methods.'' & 20 \\
\addlinespace
CatBoost\newline (1810.11363) & Gradient-boosting library for categorical features & Multi-agent multimodal accessibility & ``Gradient-boosting library for categorical features; only tangential overlap in generic ML methods.'' & 20 \\
\addlinespace
Adam\newline (1412.6980) & Stochastic optimization for deep learning & Orthogonal finetuning with matrix-free computation & ``Generic stochastic optimization method; does not address orthogonal finetuning or matrix-free computation.'' & 20 \\
\addlinespace
ResNet\newline (1512.03385) & Deep residual networks for image classification & Diffusion inference-time alignment via tree search & ``ResNet for image classification; does not address diffusion inference-time alignment or tree-search sampling.'' & 20 \\
\addlinespace
GELU\newline (1606.08415) & Activation function for neural networks & Diffusion inference-time alignment via tree search & ``Introduces an activation function; only a peripheral connection through the broader deep-learning context.'' & 20 \\
\addlinespace
Sentence-BERT\newline (1908.10084) & Sentence embedding methods & Evaluation of LLM-generated research ideas & ``Sentence embedding methods; only a tangential connection to LLM ideation-execution evaluation.'' & 20 \\
\addlinespace
PyTorch\newline (1912.01703) & Deep-learning library implementation & LLM safety and long-horizon risk simulation & ``Deep-learning library implementation; does not address LLM safety, alignment, or long-horizon risk simulation.'' & 20 \\
\addlinespace
scikit-learn\newline (1201.0490) & General-purpose machine-learning library & Hierarchical concept-bottleneck reasoning & ``General-purpose ML library; does not address hierarchical concept reasoning or graph-based CBMs.'' & 20 \\
\addlinespace
ImageNet / ILSVRC\newline (1409.0575) & Large-scale image classification benchmark & Surgical phase recognition with state-space models & ``Generic image classification benchmark; does not address surgical phase recognition or video modeling.'' & 20 \\
\addlinespace
DQN (Atari)\newline (1312.5602) & Value-based deep reinforcement learning & Hierarchical planning for long-horizon goal-conditioned tasks & ``Value-based deep RL on Atari; does not address hierarchical planning or long-horizon goal-conditioned tasks.'' & 20 \\
\bottomrule
\end{tabular}
\caption{Example low-scoring human references with the judge's abridged rationale. Score is on the 0--100 rubric. All examples drawn from the 250-paper human ground-truth set used in Section~\ref{sec:human-analysis}.}
\label{tab:qualitative_examples_appendix}
\end{table*}

\section{Artifact Licenses and Use}
\label{sec:appendix-licenses}
We use the following pre-existing artifacts in their published configurations: \textsc{GPT-OSS-120B} (Apache 2.0), \textsc{Qwen3-Embedding-8B} \citep{qwen3embedding2025} (Apache 2.0), \textsc{vLLM} \citep{kwon2023vllm} (Apache 2.0), and \textsc{HDBSCAN} \citep{mcinnes2017hdbscan} (BSD-3). Reference data comes from OpenAlex \citep{priem2022openalex} (CC0), Semantic Scholar \citep{kinney2023semanticscholar} (standard API terms of service), and arXiv preprints (per-author licenses selected at submission). All uses align with the research purposes the source artifacts permit. The artifacts we will release (the Deep Research pipeline code, the LLM-judge prompts, the OpenAlex distance-analysis scripts, and \textsc{RollingEval-Jun25}) will be made available under Apache 2.0 for code and CC-BY-4.0 for the benchmark.

\end{document}